\documentclass[10pt,twocolumn,letterpaper]{article} 

\usepackage{avss}
\usepackage{times}
\usepackage{epsfig}
\usepackage{graphicx}
\usepackage{amsmath}
\usepackage{amssymb}
\usepackage{tipa}
\usepackage{svg}
\usepackage{booktabs}
\usepackage{subfig}

\PassOptionsToPackage{hyphens}{url}\usepackage[pagebackref=true,breaklinks=true,letterpaper=true,colorlinks,bookmarks=false]{hyperref}

\avssfinalcopy 

\def\avssPaperID{2} 

\ifavssfinal\fi
\begin{document}

\title{What goes around comes around: Cycle-Consistency-based\\Short-Term Motion Prediction for Anomaly Detection using Generative~Adversarial~Networks}
\author{
Thomas Golda\textsuperscript{1,2,*}\quad
Nils Murzyn\textsuperscript{2,*}\quad
Chengchao Qu\textsuperscript{2,*}\quad
Kristian Kroschel\textsuperscript{2,*}\\
\begin{minipage}[t]{\linewidth}
    \vspace{0.001cm}
    \centering\textsuperscript{*}Fraunhofer Center for Machine Learning
\end{minipage}\\
\begin{minipage}[t]{0.5\linewidth}
    \vspace{.001cm}
    \begin{center}
    \textsuperscript{1}Vision and Fusion Lab\\
    Karlsruhe Institute of Technology KIT\\
    c/o Technologiefabrik, Haid-und-Neu-Strasse 7\\
    76131 Karlsruhe, Germany
    \end{center}
\end{minipage}
\begin{minipage}[t]{0.5\linewidth}
    \vspace{.005cm}
    \begin{center}
    \textsuperscript{2}Fraunhofer Institute for Optronics, System\\Technologies and Image Exploitation IOSB\\
    Fraunhoferstrasse 1\\
    76131 Karlsruhe, Germany
    \end{center}
    \vspace{.05cm}
\end{minipage}\\
{\tt\small firstname.lastname@iosb.fraunhofer.de}\\
}

\maketitle
\begin{abstract}
Anomaly detection plays in many fields of research, along with the strongly related task of outlier detection, a very important role.
Especially within the context of the automated analysis of video material recorded by surveillance cameras, abnormal situations can be of very different nature.
For this purpose this work investigates Generative-Adversarial-Network-based methods (GAN) for anomaly detection related to surveillance applications. 
The focus is on the usage of static camera setups, since this kind of camera is one of the most often used and belongs to the lower price segment.
In order to address this task, multiple subtasks are evaluated, including the influence of existing optical flow methods for the incorporation of short-term temporal information, different forms of network setups and losses for GANs, and the use of morphological operations for further performance improvement.
With these extension we achieved up to 2.4\% better results.
Furthermore, the final method reduced the anomaly detection error for GAN-based methods by about 42.8\%.
\end{abstract}
\thispagestyle{empty}

\section{Introduction}
With the growing number of CCTV cameras located in urban environments, the amount of recorded high-resolution data increases steadily.
A major challenge that comes to light with this is the rising difficulty of handling such large amount of data.
Despite the network traffic and storage space, the number of security staff has to be increased, since it gets almost impossible for a single person to keep an eye on every single camera. 
This leads to the demand for automatic systems that assist people in such data intensive situations.
However, most systems that deliver desired information like certain recognized activities or the detection of abandoned objects come with a very focused task. 
\begin{figure}[t]
    \begin{center}
        \includegraphics[width=0.48\linewidth]{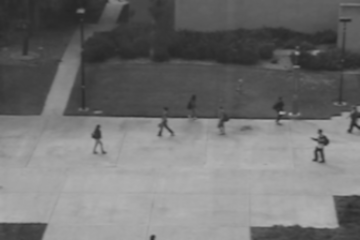}
        \hfill
        \includegraphics[width=0.48\linewidth]{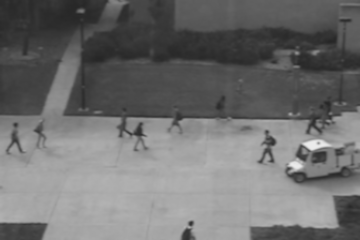}
    \end{center}
    \caption{\textbf{Examples for a normal and abnormal situation represented in the Ped2~\cite{Maha10} dataset}. The image on the left-hand side is considered as normal since only pedestrians occur. On the right-hand side the occurrence of a vehicle is considered as anomaly.}
    \label{fig:dataex}
\end{figure}
Hence, labeled data is needed, which is not available in many cases and strongly depends on the task.
Furthermore, the actual information in which security personnel is interested, differs strongly depending on the observed scenery. 
One way to address this need is to use Generative Adversarial Networks~(GANs) to learn the data distribution.
GANs are a specific neural network setup which consists of two networks: a generator network and a discriminator network.
These networks are used to capture real data distributions, for which the data can be obtained with very low effort and no need of labels or further supervision.
In order to tackle the task of unsupervised anomaly detection in videos, we follow the idea of applying GANs.
These GANs aim to perform a domain transfer between appearance information and motion information.
Based on this, our model is trained to identify the distribution of normal motion patterns and appearances and hence to distinguish those from abnormal situations.\\
Thus, in this work we evaluate several aspects of state-of-the-art anomaly detection methods with respect to the characteristics of static surveillance cameras.
Specifically, we focus on two distinct aspects in the pipeline of modern anomaly detection methods.
First, for generating training data we evaluate the impact of optical flow methods regarding the overall dynamics between successive frames.
Second, since our aim is to generate a prediction for short-term motion and compare it with the actual perceived motion, we attempt to reduce occurring noise. 
To do so we extend our baseline at two points: 
During training we include an additional cycle consistency loss for a single-frame-based motion prediction leading to less noisy anomaly heat maps.
At test time, we apply morphological operations to further improve the quality of the predicted anomaly maps.
The evaluation on two distinct GAN setups emphasizes the positive effect of our extensions.\\
This work is divided as follows:
Starting with this introduction, this work then gives a short overview on anomaly detection for video surveillance focusing on existing GAN-based methods.
The subsequent part concentrates on our experiments, which include various considered methods for optical flow computation and a cross-domain approach for training a specific GAN architecture.
Finally, the described extensions to our baseline are evaluated separately in an ablation study.

\section{Related Work}

\subsection{Overview of Anomaly Detection Methods}
According to~\cite{Kiran18} deep learning based anomaly detection methods can be divided into three different categories: representation learning for reconstruction, predictive modeling, and generative models. 
Methods categorized as representation learning for reconstruction are used to find a transformation of the training data which defines the normal behavior. 
Anomalies do not fit the implicit assumption and thus are reconstructed poorly. 
These methods are often based on auto encoders~(AEs)~\cite{XU17} or convolutional auto encoders~(CAEs)~\cite{Hasan16}. 
Predictive modeling methods aim to compute the current data sample based on the previous frames and thus have a stronger focus on temporal dependencies.
These temporal dependencies are assumed to differ greatly between normal and abnormal samples.
In order to capture the temporal dependencies of normal samples, the concepts of long short-term memory~(LSTM)~\cite{Medel16} or slow feature analysis~(SFA) \cite{Hu16} are applied.
Generative models are utilized to capture the data distribution of normal data samples.
Based on these models anomalies are detected.
Generative adversarial nets which we utilize for anomaly detection can be assigned to the latter category.

\subsection{Anomaly Detection Using GANs}\label{relatedwork:ADusingGANs}
Lots of work has been published on generative models for the task of anomaly detection, especially on GANs.
This section gives an overview over work that is strongly related to our approach and applies GANs to this task.

Lee et al.~\cite{Lee18} combined a GAN and an LSTM to a so called STAN.
Their method consists of a spatio-temporal generator and discriminator.
The generator can be subdivided into three parts:
a spatial encoder, a bidirectional convolutional LSTM (ConvLSTM) network and a spatial decoder.
The spatial encoder extracts features of an input frame.
Given a frame $F_t$, the features of the five preceding frames $F_{t-5}, ..., F_{t-1}$ and the five subsequent frames $F_{t+1}, ..., F_{t+5}$ are fed to a ConvLSTM which extracts temporal features.
A spatial decoder infers the inter-frame $\widetilde{F}_t$ based on the output of the LSTM. 
The anomaly score is generated as a weighted sum of the mean squared error between $F_t$ and $\widetilde{F}_t$, and the output of the discriminator.

Liu et al.~\cite{Liu18} applied a GAN setup to generate future frames from a sequence of input frames.
Therefore, the authors proposed to stack $t$ consecutive frames to an input data sample.
The generator used in this setup consists of several convolutional layers and learns to predict the future frame on the basis of a given set of frames.
In addition to the GAN-loss and the pixel-wise reduction loss as used in STAN, the authors introduced a frame gradient loss~\cite{Mathieu15} between a predicted frame $\widetilde{F}$ and its target frame $F$.
The used flow maps are calculated for two image pairs, the last frame $F_{t}$ of the input sequence and its generated prediction $\widetilde{F}_{t+1}$, as well as $F_{t}$ and the consecutive real frame $F_{t+1}$.
Finally, the discriminator identifies whether the input sequence of $t$ frames is part of the data distribution and hence shows a normal or abnormal situation.

\begin{figure*}[t]
    \centering\includegraphics[width=0.8\linewidth]{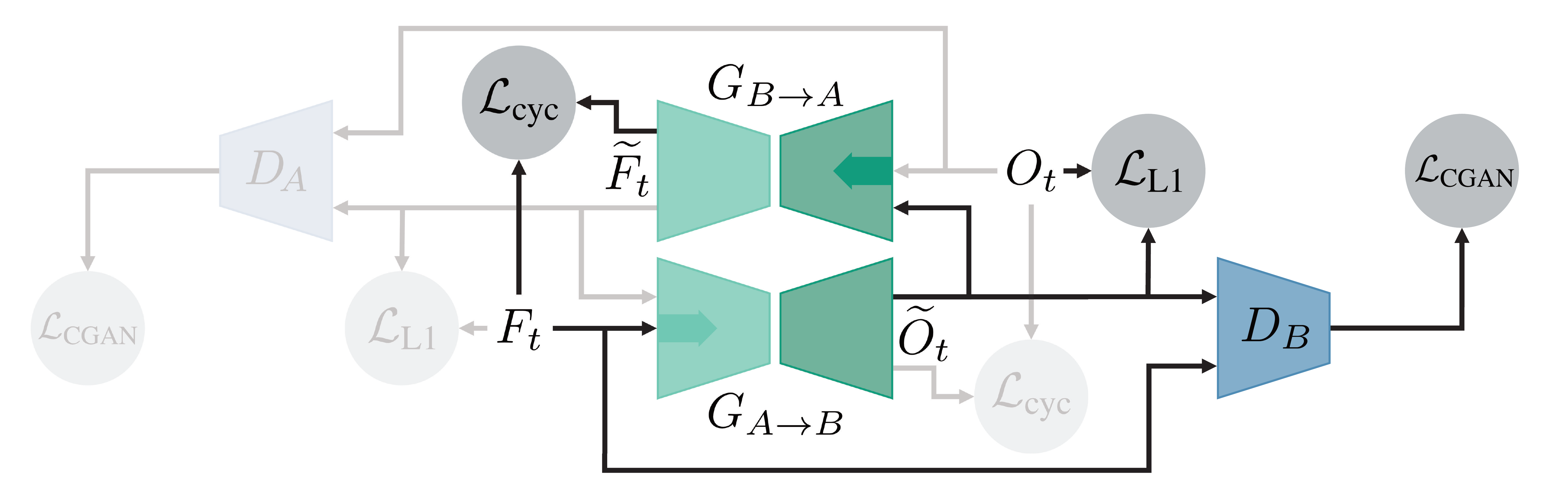}
    \vspace{-0.25cm}
    \caption{\textbf{Schematic overview of our GAN training.} During the training process of our GAN approach two generators are trained. This schematic highlights the direction from appearance to motion, where the generators $G_{A\rightarrow B}$ and $G_{B\rightarrow A}$ are trained one for each transformation direction between motion and appearance domain. The procedure for the direction from motion to appearance (see faded part) is the same.\label{fig:training}}
\end{figure*}

Ravanbakhsh et al. \cite{Ravan17b} presented a so called \emph{cross-channel approach}, trying to use two channels: motion and appearance.
Therefore, the motion channel represented by an optical flow field is predicted based on the current input frame.
At the same time, the corresponding frame of the appearance channel is reconstructed based on classical optical flow computation between the frame at time steps $t$ and $t+1$.
These corresponding generation tasks are realized using GANs.
During the training of the first GAN setup, the conditional discriminator distinguishes pairs which consist of an input frame and the optical flow computed by the generator from the input frame and its conventionally calculated optical flow. 
In the second setup, the discriminator distinguishes between pairs where the frames are either generated or real.
During inference, the discriminator is fed only with real frames and real optical flow maps. 
Normal data is part of the real data distribution and thus the discriminator assigns a high probability to a seen sample.
In contrast, abnormal data is assumed not to fit the data distribution and thus the discriminator assigns a low probability to these samples.
The outputs of the discriminators are then used to detect anomalies.

In another approach, Ravanbakhsh et al.~\cite{Ravan17a} proposed to use the same training setup but instead of using the discriminators for detection they adapted the output of the generators.
They use the difference between the generated sample and the ground truth for each channel to determine a measure for the abnormality.
In order to create a semantic difference, they applied a pretrained AlexNet~\cite{Kriz12} as feature extractor and computed the difference in the feature space. 

We will adapt this idea for our method, which will be presented in the next section.
\section{Method}\label{methods}
This section gives an overview over our studies for image based anomaly detection in surveillance scenarios.
Therefore, we adapted the cross-channel approach proposed in~\cite{Ravan17a} and applied further extensions including cycle-consistency and morphological operations to the original implementation.
In the following we present the procedure to train our GAN setup focusing on the direction from appearance domain $A$ to motion domain $B$.
The training of the opposite direction which does the transfer from motion domain to appearance domain is performed analogous.

\subsection{Cross-Channel Approach}\label{methods:cca}
We utilize camera frames and optical flow maps to realize a transfer between motion and appearance information. 
The cross-channel approach is based on the pix2pixGAN~\cite{Isola2017} which is based on a Conditional Generative Adversarial Network (CGAN) to transfer images from a source to a target domain.
In order to realize an image domain transfer, optical flow maps are interpreted as 3-channel images.
CGANs consist of a generator $G$ and a discriminator $D$, where the generator aims to capture the real data distribution.
The discriminator's task is to distinguish between real and generated data samples.
The conditional generator and the conditional discriminator get the source image $a\in A$ as additional information.
The generator aims to compute the corresponding sample $b\in B$ in the target domain as output using random noise $c$.
The loss $\mathcal{L}_{\text{CGAN}}^{\text{(V)}}(G,D)$ as introduced for the VanillaGAN~\cite{Goodf14} is given by:
\begin{equation}\label{eq:gancc}
    \begin{aligned}
	\mathcal{L}_{\text{CGAN}}^{\text{(V)}}(G,D) = \, & \mathbb{E}_{b,a}[\log D ( b\mid a)] \\
	                                            + \, & \mathbb{E}_{c,a}[1 - \log D(G( c\mid a))]
    \end{aligned}
\end{equation}
In addition to that, we performed further experiments using the GAN loss based on the least squared error~\cite{Mao17} and refer to it in the following as LSGAN.
In this respect, the loss function is split into a loss for the generator $\mathcal{L}_{\text{CGAN}}^{\text{(LS)}}(G)$ and a loss for the discriminator $\mathcal{L}_{\text{CGAN}}^{\text{(LS)}}(D)$:

\begin{figure*}[ht]
    \centering\includegraphics[width=0.8\textwidth]{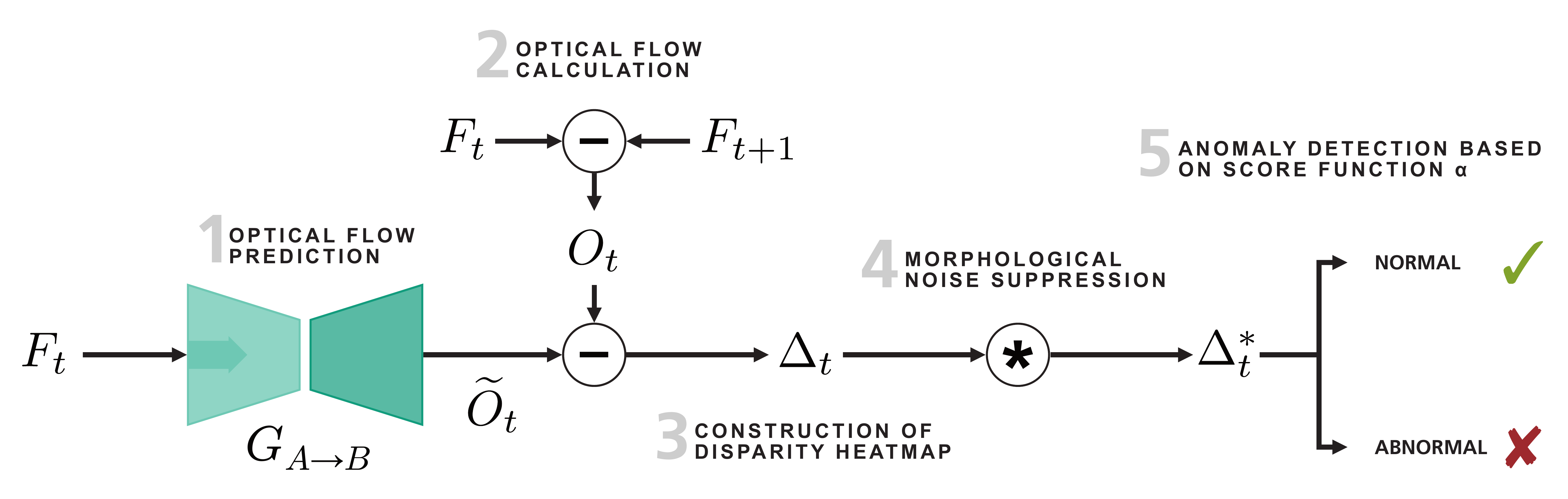}
    \vspace{-0.25cm}
    \caption{\textbf{Schematic of the anomaly detection workflow during inference.} Given a video frame $F_t$, the generator predicts dense optical flow $\widetilde{O}_t$ which is compared to the actual optical flow $O_t$. The result is a heat map $\Delta_t$. After some processing (cf. Section~\ref{sec:noisesuppression}) $\Delta_t$ is refined in order to eliminate interfering noise. Based on the resulting value of the anomaly score function $\alpha(\Delta^*_t)$ the frame $F_t$ is finally categorized as normal or abnormal.\label{fig:inference}}
    \vspace{-0.2cm}
\end{figure*}

\begin{align}
    \mathcal{L}_{\text{CGAN}}^{\text{(LS)}}(G) = \, & \frac{1}{2} \mathbb{E}_{a,c}[(D(G( c\mid a)) -1)^{2}], \\
    \begin{split}
    \mathcal{L}_{\text{CGAN}}^{\text{(LS)}}(D) = \, & \frac{1}{2} \mathbb{E}_{b,a}[(D( b\mid a) - 1 )^{2}] \\
	                                           + \, & \frac{1}{2} \mathbb{E}_{a,c}[(D(G( c\mid a)))^{2}]
    \end{split}
\end{align}

Additionally, the GAN task is extended by adding the L1-distance between the generated output and the image $b$ in the target domain to the loss function.
This loss is calculated according to:
\begin{equation}\label{eq:L1cc}
    \begin{aligned}
	\mathcal{L}_{\text{L1}}(G)=\mathbb{E}_{ b, a, c}[ \parallel G( c\mid a)- b\parallel_{1}]
	\end{aligned}
\end{equation} 
These losses are added up in a weighted sum as introduced in the original pix2pix GAN implementation~\cite{Isola2017}.
The whole setup is used in order to train generators which are able to transfer normal camera scenes in normal optical flow maps and vice versa.\footnote{The GAN is trained in both directions in an alternating manner.}
In case of transferring appearance to motion information, the optical flow map $\widetilde{O}_{t}$ is generated based on the corresponding frame $F_{t}$.
The ground truth flow map $O_{t}$ is computed based on the frame $F_{t}$ and $F_{t+1}$.
During application, abnormal scenes, i.e. scenes that were not represented in the training data, are transferred poorly which leads to a higher disparity between the target image and the generated one.
The aforementioned disparity is then used to calculate an anomaly score.
The general workflow can be described as follows: 
Firstly, for comparing optical flow maps the frame-wise anomaly score is calculated directly based on the disparity between the translated and the original flow map.
Secondly, different to~\cite{Ravan17a} we utilize the third layer of the third convolutional block (3-3) of a pretrained VGG-16 network~\cite{Simo15:VGG} to extract a flattened feature vector from video frames in order to determine a semantic difference between two frames.
Finally, the disparity between the feature map of the original and the transferred image is used as anomaly score.
In both cases the differences of the feature maps are calculated element-wise, squared and afterwards summed up along the channel dimension.
The result of the corresponding operation is a heat map $\Delta \in \mathbb{R}^{m \times n}$ of the squared differences where $m$ and $n$ denote the spatial dimensions.
These heat maps are utilized to calculate the frame wise anomaly score function $\alpha: \mathbb{R}^{m \times n} \rightarrow \mathbb{R}^{+}_{0}$ as~follows:
\begin{equation}
	\alpha(\Delta)=\sqrt{\frac{1}{m \, n} \sum_{i=1}^{m}\sum_{j=1}^{n}\Delta_{ij}}
\end{equation}
Here, $\Delta_{ij}$ denotes the entry in the \emph{i}-th row and \emph{j}-th column of the heat map $\Delta$.
Thus, the anomaly score is based on the root mean squared error between the original and the transferred representation in each domain.
As in other work~\cite{Lee18,Liu18,Ravan17a,Ravan17b}, anomaly scores are normalized video-wise. 
Since for each sample two heat maps based on the different domains are obtained, fusion as proposed in~\cite{Ravan17a} was applied. 
The fused heat map $\Bar{\Delta}_F=\Bar{\Delta}_C+\lambda_{h}\Bar{\Delta}_O$ is the result of a weighted sum of the video-wise normalized heat maps obtained from translation from camera frame to optical flow $\Bar{\Delta}_O$ and from optical flow to camera frame $\Bar{\Delta}_C$.

\begin{table*}[t!]
\caption{\textbf{Evaluation of VGG-16 conv layers for semantic difference}. The table shows the results for the anomaly detection task when using different layers of VGG-16. Conv (3-3) showed the best results for the computation of semantic differences. For this evaluation we used the second generator $G_{B\rightarrow A}$, which transforms flow to frames. This task is dominated by ambiguities between person appearance that lead to lower AUC scores.}\label{tab:vgg16}
\centering\begin{tabular}{l|cc|cc|ccc|ccc|ccc}
\toprule
\textbf{conv layer} & (1-1)   & (1-2)   & (2-1)   & (2-2)   & (3-1)   & (3-2)   & (3-3)  & (4-1)  & (4-2)  & (4-3)  & (5-1)  & (5-2)  & (5-3)  \\
\midrule
\textbf{AUC} [\%] & 65.4  & 64.9  & 62.8  & 58.6  & 63.6  & 74.3  & 78.5 & 68.0 & 66.1 & 65.0 & 57.0 & 64.6 & 64.8 \\
\bottomrule
\end{tabular}
\end{table*}
\subsection{Cycle-Consistency Extension}
For our experiments we extend the adapted GAN model by cycle-consistency as proposed in~\cite{Zhu17}.
A given sample $a$ taken from a source domain $A$ is translated to a target domain $B$ using the generator $G_{A \rightarrow B}$.
The counterpart generator $G_{B\rightarrow A}$ of the cross-channel approach aims to reconstruct the translated input sample in its original domain.
The cycle-consistency loss $\mathcal{L}_{\text{cyc}}$ is the pixel-wise L1 distance between the input and its reconstruction:
\begin{equation}\label{eq:cycleconcistency}
    \mathcal{L}_{\text{cyc}} =  \mathbb{E}_{ a, c}[\parallel G_{B \rightarrow A}( c \mid G_{A \rightarrow B}( c \mid  a))- a\parallel_{1}]
\end{equation} 
where $c$ represents random noise, incorporated by dropout within the network as introduced in~\cite{Isola2017}.
The weighted cycle-consistency loss is added to the loss for the domain transfer as introduced in Section~\ref{methods:cca}.
Substituting the corresponding generators in the loss terms leads to the composite objectives for VanillaGAN and~LSGAN shown in Eq.~\ref{eq:vanillagan}, \ref{eq:lsgan_g} and~\ref{eq:lsgan_d}. The overall schematic is shown in Fig.~\ref{fig:training}.
\begin{align}
    \begin{split}
    \mathcal{L}^{(V)} & = \mathcal{L}_{\text{CGAN}}^{(V)}(G_{A \rightarrow B},D_B) \\
                      & + \lambda_{\text{L1}}\mathcal{L}_{\text{L1}}(G_{A \rightarrow B}) \\
                      & + \lambda_{\text{cyc}}\mathcal{L}_{\text{cyc}}
    \end{split}\label{eq:vanillagan}
    \\[2ex]
    \begin{split}
    \mathcal{L}^{(LS)}_G & = \mathcal{L}_{\text{CGAN}}^{(LS)}(G_{A \rightarrow B}) \\
                       & + \lambda_{\text{L1}}\mathcal{L}_{\text{L1}}(G_{A \rightarrow B}) \\
                       & + \lambda_{\text{cyc}}\mathcal{L}_{\text{cyc}}
    \end{split}\label{eq:lsgan_g}
    \\[2ex]
    \mathcal{L}^{(LS)}_D & = \mathcal{L}_{\text{CGAN}}^{(LS)}(D_B)\label{eq:lsgan_d}
\end{align}

\subsection{Noise Suppression}\label{sec:noisesuppression}
Under the assumption that anomalies in surveillance cause spatially larger areas of differences in the heat map, during application phase we perform in a post-processing step the suppression of small area differences which are considered as noise.
We applied classical morphological operations, namely \emph{closing} and \emph{opening}, to achieve the desired behavior.
In order to do so, $\Delta$ has to be interpreted as a binary image.
This is achieved by clipping the values of each bin $\Delta_{ij} > 0$ to $1$.
Closing is then applied to eliminate small holes from large area segments, which makes large area differences robust to opening which is applied afterwords to eliminate the small area differences and thus reduce false positive predictions.
The resulting refined heat map is denoted as $\Delta^*$.
For an overall overview of the workflow during inference see Fig.~\ref{fig:inference}.
\section{Experiments}\label{experiments}

\begin{figure*}[t]
    \begin{center}
        \includegraphics[width=0.32\linewidth]{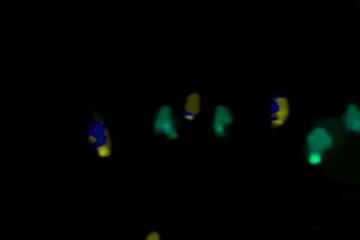}
        \hfill
        \includegraphics[width=0.32\linewidth]{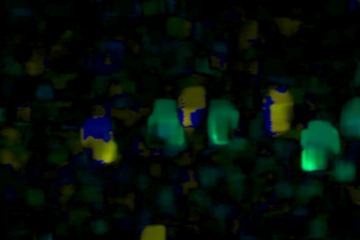}
        \hfill
        \includegraphics[width=0.32\linewidth]{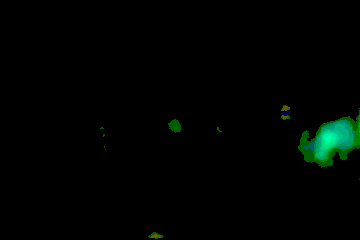}
    \end{center}
   \caption{\textbf{Visualized flow maps of different calculation methods}. The flow map calculated according to~\cite{Brox04:clasOptFlow} is visualized on the left-hand side, the one according to~\cite{Farn03:optFlow} in the middle and the one according to~\cite{Ilg17:FN2} on the right-hand side. Color intensities were adjusted to improve the visualization. The hue indicates the direction of the movement. The corresponding camera frame is visualized in Fig.~\ref{fig:dataex} on the left-hand side. As can be seen, there is a clear difference between the quality of the optical flow maps, where the left one delivers the clearest and best concentrated flow maps compare to the middle and right one.}\label{fig:flowcomp}
\end{figure*}


\subsection{Dataset}
For the evaluation, we performed the experiments on the UCSD dataset~\cite{Maha10}.
The UCSD data set is split into two subsets: Ped1 and Ped2.
Each subset shows a university walkway in which pedestrians are considered as normal and unusual objects such as vehicles and atypical motion patterns such as cycling or skating are considered anomalous.
Furthermore, all frames within the same subset show the identical background.
The Ped1 data set is composed of 34 training videos and 36 testing videos.
Each video comprises 200 frames at a resolution of $238 \times 158$ pixels.
The total number of frames is 14,000 with 40 different abnormal events.
The Ped2 data set contains 16 training videos and 12 testing videos.
The number of frames per video ranges between 120 and 180, summing up to 4,560 frames in total with 12 different types of anomalies.
The resolution is $360 \times 240$ pixels and thus slightly higher than in Ped1.
Fig.~\ref{fig:dataex} shows an example of a normal scene and an abnormal scene of the Ped2 data set.
According to the number of citations, this data set is the most used benchmark for anomaly detection in surveillance.
However, as some recent work reports results for Ped1 on a subset of 16 test videos~\cite{Ravan17a} whereas others report results for 36 videos~\cite{Liu18}, we decided to conduct our experiments on the Ped2 subset.
The metric used for quantitative comparison is the area under the receiving operating characteristic (AUC).

\subsection{Implementation Details}
All camera images and optical flow maps were scaled to a resolution of $256 \times 256$ pixels.
To visualize the flow map as a three channel image the angle and the magnitude of the flow vector were interpreted as the hue and the intensity of the HSI color space.
The saturation was set to $1$ in order to improve clarity.
The intensities of the grayscale camera images are copied to each RGB channel.

The generator was realized as a U-Net generator and the discriminator was realized as a PatchGAN discriminator as introduced in the original set up~\cite{Isola2017}.
The GAN was trained for 10 epochs utilizing the Adam optimizer~\cite{Kingma14:Adam}.
Each translation direction of the baseline model was trained independently, whereas both translation directions of the model with cycle-consistency extension are trained simultaneously to improve stability~\cite{Zhu17}. 
The morphological operations apply a structural element with a kernel size of $7 \times 7$ in which each element has a value of 1 in order to perform the noise suppression on areas containing differences. 
The stride is set to 1.

\subsection{Impact of Optical Flow Calculation}

In order to determine the impact of the optical flow calculation method, we investigated three different methods to calculate optical flow according to~\cite{Brox04:clasOptFlow}, \cite{Farn03:optFlow} and~\cite{Ilg17:FN2}.
The methods proposed in \cite{Brox04:clasOptFlow} and \cite{Farn03:optFlow} are classic methods whereas the method in \cite{Ilg17:FN2} also known as FlowNet2 is based on neural networks.
In Fig.~\ref{fig:flowcomp} the flow maps of the different calculation methods are depicted.

\begin{table}[h]
	\centering
	\caption{\textbf{Evaluation of different methods for optical flow generation.} The table shows AUC results  using the generator $G_{A\rightarrow B}$ (flow-based), $G_{B\rightarrow A}$ (frame-based), and fused results for our VanillaGAN using optical flow generated according to \cite{Brox04:clasOptFlow}, \cite{Farn03:optFlow} and \cite{Ilg17:FN2}. The best results were achieved using the method proposed in \cite{Brox04:clasOptFlow} in all cases.}
	\begin{tabular}{c c c c} 
	    \toprule
	                                    & \multicolumn{3}{c}{\bfseries Ped2}\\
        \cmidrule{2-4}
		\bfseries Method                & \bfseries frame        & \bfseries fused  & \bfseries flow\\ 
		\midrule
		Brox \cite{Brox04:clasOptFlow}  & \textbf{78.5\%}        & \textbf{85.8\%}  & \textbf{93.7\%}\\ 
		Farneback \cite{Farn03:optFlow} & 69.6\%                 & 80.8\%           & 84.2\% \\ 
		FlowNet2 \cite{Ilg17:FN2}       & 66.0\%                 & 75.0\%           & 81.5\% \\ 
		\bottomrule
	\end{tabular}
	\label{tab:AUCgt}
\end{table}

As evident from the comparison of AUCs (area under curve) in Tab.~\ref{tab:AUCgt}, calculations according to \cite{Brox04:clasOptFlow} achieved best performance with respect to anomaly detection.
We identify the reason for this performance in the high frequency component calculation in the optical flow maps.
As visualized in Fig.~\ref{fig:flowcomp} the method according to \cite{Brox04:clasOptFlow} calculates higher frequency components and discontinuities within the optical flow field.
Due to its assumption of a slowly varying flow field, the method according to \cite{Farn03:optFlow} is well known to smooth out discontinuities.
These components though are important to characterize the motion patterns in greater detail and hence differ between different motion patterns such as those of pedestrians and skateboarders.
Consistently, anomaly detection is sensitive to the optical flow calculation method based on which motion patterns are captured.
The optical flow maps calculated with the pretrained FlowNet2~\cite{Ilg17:FN2} visualize motion patterns poorly.
An explanation for these poor calculations is that the data on which this network is pretrained does not cover scenes with static backgrounds and multiple, similarly moving objects.


Further, the results imply that the translation from camera images to optical flow achieves better performance.
The main reasons for the poor performance of the detection by translation from optical flow to frame is the difficult reconstruction of object attributes and backgrounds, since this information is not explicitly provided by optical flow maps and thus implicit assumptions by the generator are needed.
These implicit assumptions though do not have any effect on the anomaly detection task.
Fusion of heat maps which are computed based on the translation in the two different directions as proposed in \cite{Ravan17a} builds approximately the mean of both receiving operating characteristics and thus does not yield a further improvement.
This is in contrast to findings in \cite{Ravan17a} which could be caused by the utilization of a maximum difference based anomaly score calculation method. 
In general, we observed that heat maps computed by the semantic comparison of camera frames contain large areas of small difference for normal objects which can be interpreted as noise. 
This noise achieves it maximal value in the center of these normal objects. 
The noise caused in heat maps computed based on the comparison of optical flow maps is located at the extremities of pedestrians whereas differences caused by an anomaly are distributed over the entire object shape.
In case of a maximal difference based calculation method, the areas of maximal differences caused by noise do not superpose after fusion and devalue these differences relative to differences caused by anomalies. 
Thus, fusion smooths out the high values of the wrongly predicted extremity movement of pedestrians in the flow domain. 
In case of a root mean squared differences based anomaly score calculation the impact of the in general larger areas of differences in heat maps computed based on camera frames dominates and causes blurriness.

\begin{figure}[!ht]
    \begin{center}
        \includegraphics[width=0.5\linewidth]{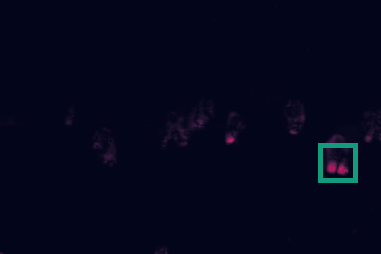}
        \includegraphics[width=0.335\linewidth]{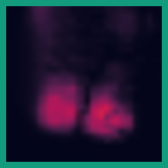}\\
        \includegraphics[width=0.5\linewidth]{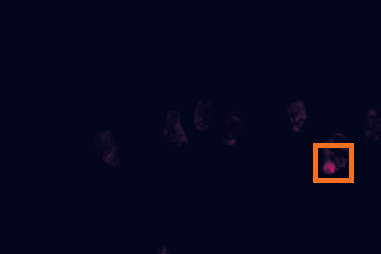}
        \includegraphics[width=0.335\linewidth]{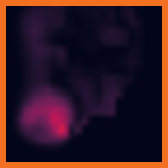}\\
        \includegraphics[width=0.5\linewidth]{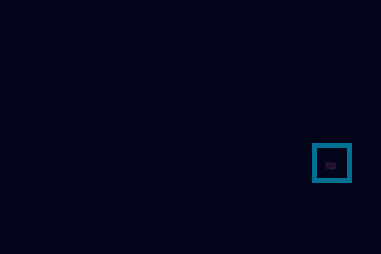}
        \includegraphics[width=0.335\linewidth]{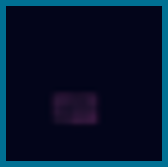}
    \end{center}
    \caption{\textbf{Heat maps for baseline and extensions.} In the first row, the heat map of the baseline model is visualized. 
    The second row contains the heat map of the same scene computed by the model with cycle-consistency extension. 
    The third row contains the heat map of the model with cycle-consistency and additional noise suppression. 
    In the left column, an area of the heat map is highlighted in which the motion of the feet of a pedestrian are by mistake classified as anomalous. 
    Cycle-consistency reduces this prediction error and the noise suppression step almost eliminates it completely.}
    \label{fig:heatmaps}
\end{figure}

\subsection{Extensions}
We applied our extensions to the better performing imate-to-optical-flow setup.
Tab.~\ref{tab:Ped2ComparisonExt} shows that the extension by cycle-consistency improved performance of the VanillaGAN baseline by 1.1\%.
The additional application of noise suppression caused a further increase of performance by additional 0.9\%.
Despite the VanillaGAN loss, we utilized the LSGAN loss, which improved the performance of the baseline by 1.7\%. 
The proposed extensions likewise improved the performance of this alternative setup. 
In this setup, the cycle-consistency extension enhanced the performance by 2.2\%.
The noise suppression caused an additional improvement by 0.4\%.
These results show that our extensions generalize to other GAN setups.

\begin{table}[h]
	\centering
	\caption{\textbf{Comparison of GAN based cross-channel anomaly detection methods for Ped2.} 
	Methods investigated in the scope of this work achieve competitive performance to state-of-the-art algorithms in terms of AUC.}
	\begin{tabular}{l c} 
	    \toprule
	    \multicolumn{1}{c}{\bfseries Method}    & \bfseries AUC \\ 
		\midrule
		VanillaGAN (baseline)                   & 93.7\% \\
		VanillaGAN (cyc.-con.)                  & 94.8\% \\
		VanillaGAN (cyc.-con. + noise supp.)    & 95.7\% \\ 
		LSGAN (baseline)                        & 95.4\% \\
		LSGAN (cyc.-con.)                       & 97.6\% \\
		LSGAN (cyc.-con. + noise supp.)         & \textbf{98.0\%} \\ 
		\midrule
		Ravanbakhsh et al. \cite{Ravan17a}      & 93.5\% \\
		\bottomrule
	\end{tabular}
	\label{tab:Ped2ComparisonExt}
\end{table}

We identify the main reason for this improvement in terms of anomaly detection performance caused by the proposed extension in the reduction of faulty motion predictions. 
This effect can be observed in Fig.~\ref{fig:heatmaps}. 
In the first row the heat map of the baseline model is visualized. 
The second row contains the heat map of the model extended by cycle-consistency. 
The third row contains the heat map after additional noise suppression. 
The right column shows a magnification of the corresponding area on the left side where the motion of the feet of a pedestrian is predicted. 
The corresponding camera frame of the scene is visualized in Fig.~\ref{fig:dataex} on the left-hand side. 
The heat map computed based on the baseline model contains large differences in this area which is caused by a large disparity between the generated motion and the ground truth. 
Thus the sample is more likely to be by mistake classified as anomalous.
The extension by cycle-consistency reduces this disparity significantly and causes a smaller area of differences. 
The noise suppression eliminates the differences almost completely and hence a small anomaly score gets assigned to this scene.
However, the results also underline that the noise suppression has no significant drawbacks for the detection of actual anomalies.
This is mainly due to the cycle-consistency loss, which already improves the motion prediction for normal situations and hence leads to less false positives in the anomaly heat map.
Since the motion in abnormal situations is still predicted poorly, the further post-processing using morphological operations mainly affects the already reduced false positives.

A comparison to the methods introduced in Section~\ref{relatedwork:ADusingGANs} with our best method is given in Tab.~\ref{tab:Ped2ComparisonLit}. 
As the results show, we surpass all other GAN-based approaches and achieve better results than the methods of \cite{Lee18} and \cite{Liu18}, which use a longer time span and thus a larger number of consecutive frames for detecting anomalies.

\begin{table}[h]
	\centering
	\caption{\textbf{AUCs of GAN based anomaly detection methods for Ped2.} 
	Methods investigated in the scope of this work outperform state-of-the-art algorithms in terms of AUC.}
	\label{tab:Ped2ComparisonLit}
	\begin{tabular}{l c} 
	    \toprule
		\multicolumn{1}{c}{\bfseries Method}        & \bfseries AUC  \\ 
		\midrule
        LSGAN (cyc.-con. + noise supp.)             & \textbf{98.0\%} \\ 
        \midrule
		Ravanbakhsh et al. \cite{Ravan17b}          & 95.5\% \\
		Ravanbakhsh et al. \cite{Ravan17a}          & 93.5\% \\
		Lee et al. \cite{Lee18}                     & 96.5\% \\
		Liu et al. \cite{Liu18}                     & 95.3\% \\
		\bottomrule
	\end{tabular}
\end{table}

\subsection{Discussion}
The strength of GANs lay in learning to transfer between low-varying distributions like those of rigid cameras.
For a static camera setup a GAN-based approach delivers good results, as shown in this work. 
Only small effort has to be made to transfer it to another setup, since only the training has to be repeated.
The processing time for each timestep is in total about 40 milliseconds, which shows the applicability for real-time scenarios.
However, one drawback of these methods is scalability since for every camera a model has to be trained.
Nevertheless, this can be done quite easily and automatically.
Furthermore, the adaption to non-static cameras like PTZ or mobile cameras is a very challenging task, since the variance within the single domains is~larger.
\section{Conclusion \& Future Work}
In this work we examined GAN-based methods for anomaly detection in the scope of static camera applications.
For this reason, we investigated the influence of three different optical flow calculation methods and demonstrated that our methods extended with cycle-consistency and noise reduction improve the performance of the selected GAN architectures by 2.0\% and 2.4\%, respectively, for VanillaGAN and LSGAN.
With our final model we even outperform other state-of-the-art methods based on GANs for the task of anomaly detection and diminish the classification error by about 42.8\%. Future work on this topic will consist of more experiments using alternative mechanisms to calculate optical flow as it is done e.g. by FlowNetSD~\cite{Ilg17:FN2} in order to increase processing speed, and the transfer to non-static cameras, which brings up new challenges.

{\small
\bibliographystyle{ieee}
\bibliography{egbib}
}


\end{document}